\documentclass{article}

\usepackage{titletoc}  %
\usepackage{microtype}
\usepackage{graphicx}
\usepackage{subcaption}  %
\usepackage{booktabs}

\usepackage[accepted]{icml2025}
\usepackage{multirow}

\definecolor{darkgreen}{rgb}{0,0.40,0}
\definecolor{firebrick}{rgb}{0.698,0.133,0.133}

\usepackage[colorlinks,citecolor=darkgreen,linkcolor=firebrick,urlcolor=firebrick,linktocpage,unicode]{hyperref}

\usepackage{amsmath}
\usepackage{amssymb}
\usepackage{mathtools}
\usepackage{amsthm}
\usepackage[capitalize,noabbrev]{cleveref}
\usepackage{tcolorbox}
\usepackage{enumitem}
\usepackage{bbm}
\usepackage{wrapfig}
\usepackage[accepted]{icml2025}

\usepackage{breakurl}

\usepackage{array} %

\definecolor{LightCyan}{rgb}{0.8, 0.9, 1}

\newlength{\Oldarrayrulewidth}

\newcommand{\cellcolorcontents}[2]{%
  \begingroup
  \setlength{\fboxsep}{0pt}%
  \colorbox{#1}{\strut #2}%
  \endgroup
}

\newcolumntype{C}[1]{>{\collectcell\cellcolorcontents{LightCyan}}c<{#1}} %

\definecolor{LightCyan}{rgb}{0.8, 0.9, 1}
\newcolumntype{b}{>{\columncolor{LightCyan}\hspace{0pt}}c}
\definecolor{cadetgrey}{rgb}{0.57, 0.64, 0.69}
\newcolumntype{g}{>{\columncolor{cadetgrey}\hspace{0pt}}c}

\setlist[itemize]{leftmargin=1em, before=\vspace{-0.5em}, after=\vspace{-0.5em}, itemsep=0.1em}
\setlist[enumerate]{leftmargin=1.4em, 
before=\vspace{-0.5em}, after=\vspace{-0.5em}, 
itemsep=0.1em}

\theoremstyle{plain}

\theoremstyle{definition}

\theoremstyle{remark}

\icmltitlerunning{Efficient Temporal Tokenization for Mobility Prediction with Large Language Models
}

\begin{document}

\twocolumn[
\icmltitle{Efficient Temporal Tokenization for Mobility Prediction with Large Language Models
}

\icmlsetsymbol{equal}{*}
\begin{icmlauthorlist}
\icmlauthor{Haoyu He}{equal,affiliation1}
\icmlauthor{Haozheng Luo}{equal,affiliation2}
\icmlauthor{Yan Chen}{affiliation2}
\icmlauthor{Qi R. Wang}{affiliation1}
\end{icmlauthorlist}

\icmlaffiliation{affiliation1}{Northeastern University, Boston, MA}
\icmlaffiliation{affiliation2}{Northwestern University, Evanston, IL}

\icmlcorrespondingauthor{Qi R. Wang}{q.wang@northeastern.edu}

\icmlkeywords{Human mobility prediction, large language models, spatio-temporal forecasting}

\vskip 0.3in
]

\newcommand{\sys}{{\rm RHYTHM }}

\newcommand{\syb}{{\rm \textbf{RHYTHM }}}

\printAffiliationsAndNotice{\icmlEqualContribution}

\begin{abstract}
We introduce \textbf{RHYTHM} (\underline{R}easoning with \underline{H}ierarchical \underline{T}emporal \underline{T}okenization for \underline{H}uman \underline{M}obility), a framework that leverages large language models (LLMs) as spatio-temporal predictors and trajectory reasoners. RHYTHM partitions trajectories into daily segments encoded as discrete tokens with hierarchical attention, capturing both daily and weekly dependencies while substantially reducing the sequence length. Token representations are enriched with pre-computed prompt embeddings via a frozen LLM, enhancing the model's ability to capture interdependencies without extensive computational overhead. By freezing the LLM backbone, RHYTHM achieves significant computational efficiency. Evaluation on three real-world datasets demonstrates a \textbf{2.4\%} improvement in accuracy, \textbf{5.0\%} increase on weekends, and \textbf{24.6\%} reduction in training time compared to state-of-the-art methods.

\end{abstract}

\vspace{-1.5em}
\section{Introduction}

We propose \syb (\underline{R}easoning with \underline{H}ierarchical Temporal \underline{T}okenization for \underline{H}uman \underline{M}obility), a novel foundation model for human mobility prediction. Our approach reconceptualizes mobility modeling through structured temporal abstraction, combining efficient multi-scale temporal tokenization with the complex reasoning capabilities of pretrained LLMs. This integration yields a computationally efficient yet powerful framework for trajectory prediction, offering superior accuracy while maintaining efficiency and scalability across diverse mobility contexts.

\begin{figure}[t]
    \centering
    \includegraphics[width=0.7\columnwidth]{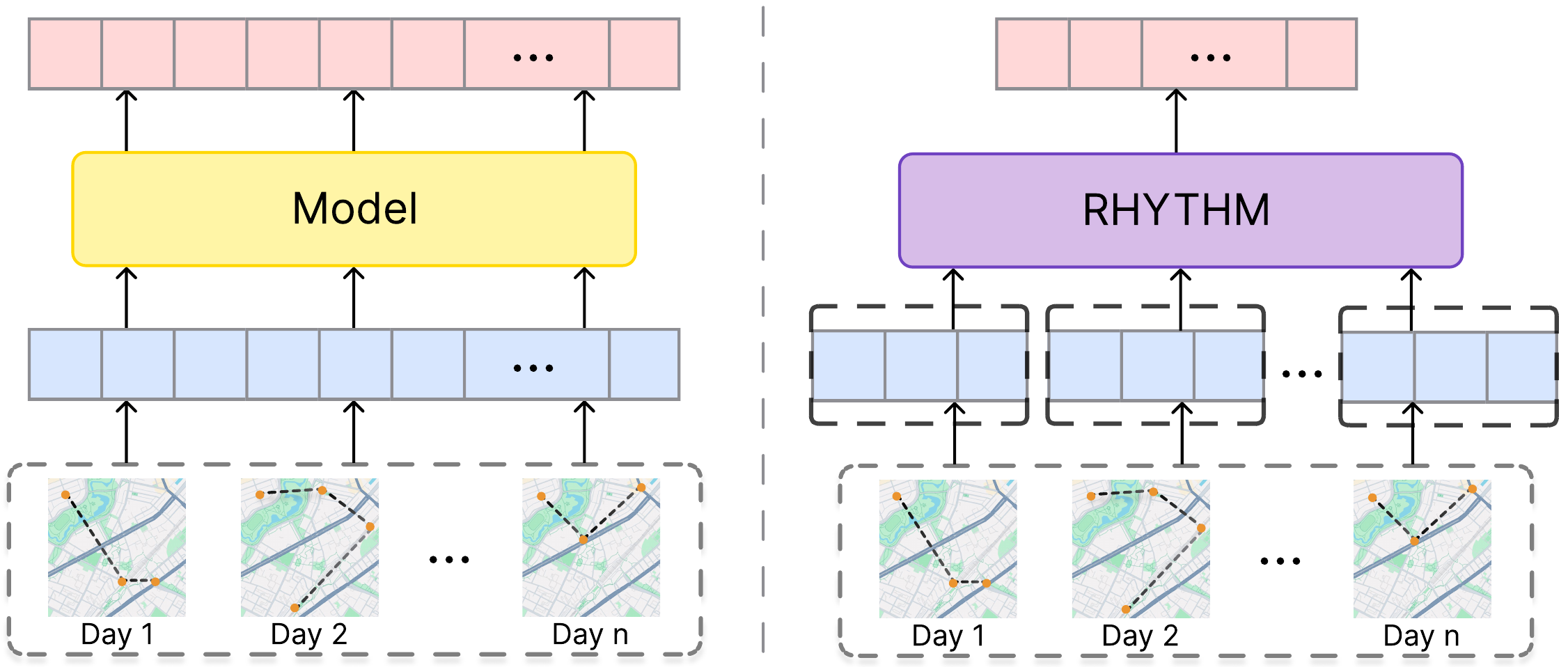}
    \vspace{-1em}
    \caption{\textbf{Motivation for RHYTHM.}  
        By partitioning trajectories into discrete tokens instead of a continuous stream, RHYTHM more effectively captures recurring mobility patterns.
    }
    \label{fig:motivation}
    \vspace{-2em}
\end{figure}

Our framework is grounded in the inherent patterns of human mobility. Research shows that human movement follows predictable daily and weekly rhythms~\cite{cho2011friendship, gonzalez2008understanding, eagle2006reality}, with \citet{song2010limits} demonstrating that 93\% of daily trajectories are predictable. Yet modeling the complex interdependence between locations and temporal cycles remains challenging. Existing approaches either neglect long-term periodicity \cite{yang2020location, feng2018deepmove, gambs2012next} or fail to capture multi-scale temporal patterns \cite{hong2023context, wu2019graph}.
We address these limitations by decomposing the trajectories into meaningful segments and tokenizing them into discrete representations. Our approach employs intra-segment attention for local patterns and inter-segment attention for long-range dependencies (see~\cref{fig:motivation}), significantly reducing computational complexity. Each token is further enriched with pre-computed prompt embeddings derived from a frozen LLM, integrating trajectory context and task descriptions to enhance semantic understanding.

Recent research highlights LLMs' dual capabilities as both representation extractors for spatio-temporal patterns and sophisticated reasoning engines~\cite{chowdhery2022palm}. These models excel through few-shot prompting~\cite{brown2020language}, chain-of-thought reasoning~\cite{pan2024chain,pan2024conv,wei2022chain}, and in-context learning~\cite{dong2022survey}. While mobility-specific models like PMT~\cite{wu2024pretrained} and ST-MoE-BERT~\cite{he2024st} lack LLM integration to model complex human movement correlations, RHYTHM addresses this limitation through an integrated LLM-based reasoning module. Our parameter-efficient strategy freezes the pre-trained LLM backbone, avoiding extensive fine-tuning while preserving reasoning capabilities. This approach effectively balances fine-grained spatio-temporal modeling with minimal computational overhead, making RHYTHM particularly suitable for resource-constrained settings.

In summary, our key contributions are:
\vspace{-0.5em}
\begin{itemize}[leftmargin=*]
  \item We introduce a novel temporal tokenization scheme that represents daily mobility routines as discrete tokens, capturing multi-scale cyclical dependencies while markedly reducing sequence length.
  \vspace{-0.5em}
  \item We propose an efficient prompt-guided integration of semantic trajectory cues and task descriptions via segment embeddings, enhancing interpretability of complex mobility behaviors.
  \vspace{-0.5em}
  \item We design parameter-efficient LLM adaptation strategy leveraging frozen backbones, requiring only \textbf{12.37\%} of model parameters and cutting computational cost by \textbf{24.6\%}.
  \vspace{-0.5em}
  \item Evaluation on three real-world datasets, yielding a \textbf{2.4\%} overall accuracy gain and a \textbf{5.0\%} improvement on weekend predictions over state-of-the-art baselines.
\end{itemize}

\begin{figure*}[htp]
    \centering
\begin{minipage}{0.69\linewidth}
    \includegraphics[width=\linewidth]{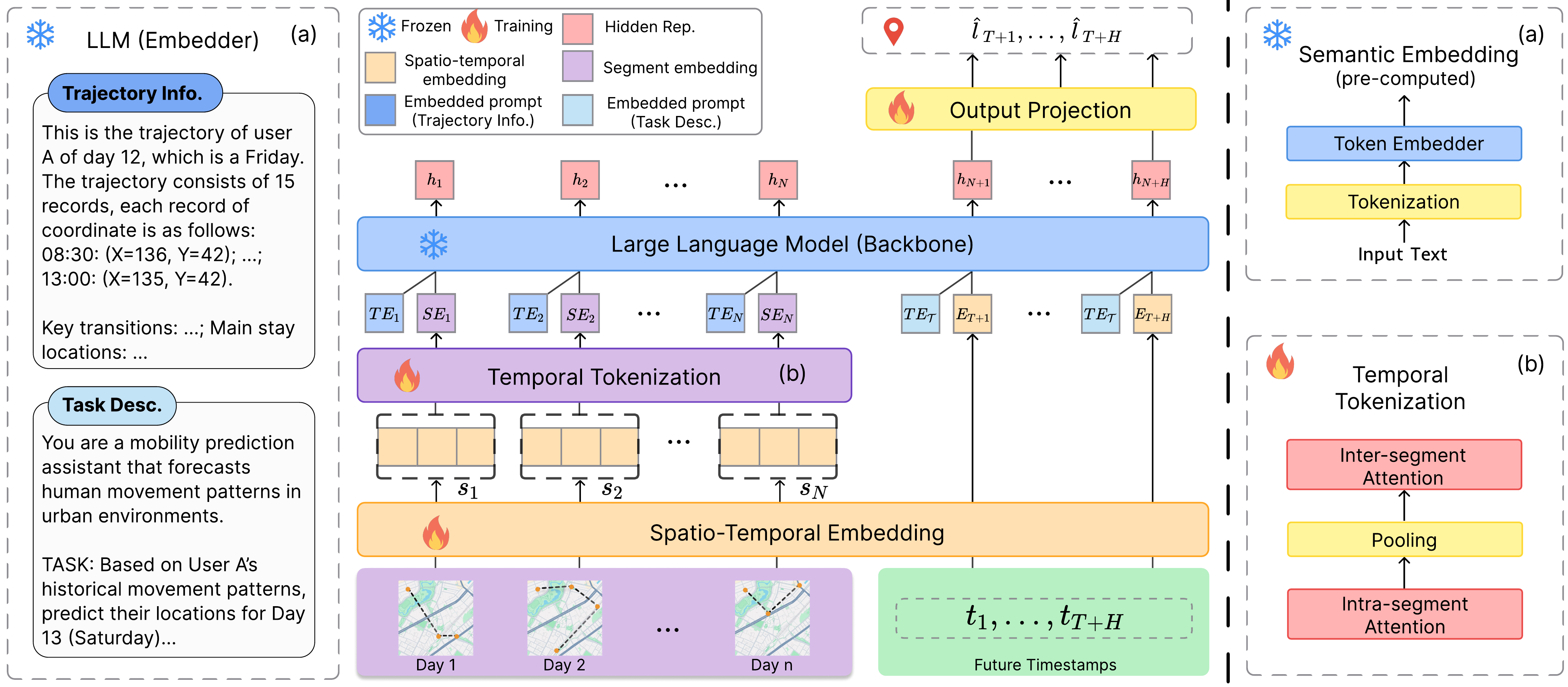}
\end{minipage}
\hfill
\begin{minipage}{0.29\linewidth}
     \caption{\textbf{The proposed architecture of RHYTHM.} Historical trajectories are first converted into spatio–temporal embeddings and discretized via temporal tokenization (\textbf{b}), enabling hierarchical attention to capture both local and global dynamics. Each segment token is enriched with semantic trajectory embeddings, while future time‐step tokens integrate task‐context descriptors (\textbf{a}). The resulting token sequence is fed into a frozen LLM backbone, and an output projection layer produces the final location predictions.
}
    \label{fig:rhythm}
\end{minipage}   
    \vspace{-0.15in}
\end{figure*}

\vspace{-1em}
\section{Method}

\label{sec:method}
\crefname{tcolorbox}{box}{boxes}

In this section, we present \textbf{RHYTHM}, an LLM-driven architecture for prompt-guided representation learning of periodic spatio-temporal patterns (see \cref{fig:rhythm}).
\vspace{-.5em}

\subsection{Problem definition}

Let $\mathcal{X} = \{x_1, x_2, \ldots, x_T\}$ be a user’s trajectory, where each observation $x_i=(t_i, l_i)$ pairs a timestamp $t_i$ with a location $l_i\in\mathcal{L}$ drawn from a finite set $\mathcal{L}$.  For a prediction horizon $H$, define the future time points $\mathcal{T}=\{t_{T+1},\ldots,t_{T+H}\}$.  Our goal is to predict the corresponding locations $\mathcal{Y}=\{l_{T+1},\ldots,l_{T+H}\}$. Formally, we learn a mapping
$
f:\;(\mathcal{X},\,\mathcal{T})\;\mapsto\;\mathcal{Y}
$
that takes the historical trajectory and future timestamps to the future location sequence.

\subsection{Model structure}

\paragraph{Spatio-Temporal Feature Encoding.}
For each observation $x_i$, we design temporal representations to encode the periodic nature of human mobility:
\begin{align*}
    \mathbf{E}^{\text{temporal}}_i = \mathbf{E}^{\text{ToD}}(t_i) \Vert \mathbf{E}^{\text{DoW}}(t_i),
\end{align*}
where $\cdot \Vert \cdot$ denotes the concatenation operation, $\mathbf{E}^{\text{ToD}}$ encodes time-of-day information, and $\mathbf{E}^{\text{DoW}}$ captures day-of-week patterns.
These trainable embeddings transform discrete time indices into dense vector representations, yielding $\mathbf{E}^{\text{temporal}}_i \in \mathbb{R}^D$ where $D$ corresponds to the input dimensionality of the underlying LLM.

The spatial representation $\mathbf{E}^{\text{spatial}}_i \in \mathbb{R}^D$ for location $l_i$ is formulated as:
\begin{equation*}
    \mathbf{E}^{\text{spatial}}_i = \mathbf{E}^{\text{Loc}}(l_i) \Vert (W_{\text{coord}}[\text{lat}_i, \text{lon}_i]^T + b_{\text{coord}}),
\end{equation*}
where $\mathbf{E}^{\text{Loc}}$ represents the location-specific categorical embedding, while the latter term transforms the geographic coordinates ($\text{lat}_i, \text{lon}_i$) to the embedding domain through the transformation matrix $W_{\text{coord}} \in \mathbb{R}^{d_{\text{coord}}\times 2}$ and $d_{\text{coord}}$ indicates the output dimension of the projection.

The unified spatio-temporal representation $\mathbf{E}_i \in \mathbb{R}^D$ is computed through element-wise summation:
\begin{equation*}
    \mathbf{E}_i = \mathbf{E}^{\text{temporal}}_i + \mathbf{E}^{\text{spatial}}_i.
\end{equation*}
When handling incomplete historical observations, we assign zero values to the spatial components, enabling the model to function solely with temporal features while maintaining consistent dimensionality.

\paragraph{Temporal Tokenization.}
Human movement behaviors demonstrate intrinsic hierarchical temporal patterns that encompasses both local behavioral routines and cyclical regularities~\cite{song2010limits,gonzalez2008understanding}.
To address these multi-scale dynamics, RHYTHM introduces a temporal segmentation strategy that separates fine-grained local behaviors from overarching temporal relationships, drawing inspiration from \citet{liu2024autotimes}.
Specifically, we divide the embedded representation $\mathcal{X}$ into $N$ distinct, consecutive segments $\{\mathbf{s}_1, \ldots, \mathbf{s}_N \}$, where each segment encodes semantically coherent time periods (such as daily sequences):
\begin{align*}
    \mathbf{s}_i = \{E_{(i-1)L+1},\dots,E_{iL}\} \quad \text{for } i = 1, \dots, N,
\end{align*}
with $L$ denoting the temporal span within each segment $\mathbf{s_i}$.

For capturing local sequential patterns within segments, we utilize intra-segment attention:
\begin{align*}
    \tilde{\mathbf{E}}^{(i)} = \mathrm{Attention}(\mathbf{s}_i),
\end{align*}
We implement a pre-normalized transformer design incorporating gated feed-forward modules following \citet{dubey2024llama3}, facilitating stable gradient propagation and enhancing representational capacity. Detailed architectural specifications can be found in~\cref{section:attention}.

To efficiently encode relationships across segments, we introduce a trainable aggregation mechanism that compresses each segment into a compact token:
\begin{align*}
    \mathbf{SE}_i = \mathrm{Pool}\bigl(\tilde{\mathbf{E}}^{(i)}\bigr).
\end{align*}
These compressed segment representations $\{\tilde{\mathbf{SE}}_1, \ldots, \tilde{\mathbf{SE}}_N \}$ are then processed through inter-segment attention to encode extended temporal relationships and distant dependencies:
\begin{align*}
    \tilde{\mathbf{SE}}_{1:N} = \operatorname{Attention}(\mathbf{SE}_{1:N}),
\end{align*}
producing enhanced segment embeddings $\mathbf{\tilde{SE}_i} \in \mathbb{R}^D$ that incorporate contextual signals from various temporal granularities.
This design compresses the operational sequence length from $T$ to $N$ while maintaining both detailed temporal features and long-range correlations, thereby mitigating the computational burden associated with processing lengthy mobility sequences.

\paragraph{Semantic Context Integration.}
The integration of LLMs into mobility prediction presents a fundamental tension: maximizing semantic expressiveness while maintaining computational tractability. Existing approaches like LLM-Mob \cite{wang2023would} adopt monolithic prompting strategies that encode entire trajectories as single, lengthy prompts—an approach that incurs prohibitive computational costs and dilutes fine-grained temporal signals. We propose a hierarchical prompting mechanism that decomposes trajectories into semantically coherent segments, each associated with a focused prompt that captures local movement patterns, transition behaviors, and activity semantics. This decomposition achieves an optimal balance: preserving the semantic richness necessary for accurate prediction while dramatically reducing the computational footprint.

Our semantic encoding pipeline operates at two granularities. For historical segments, we generate structured descriptions that capture movement patterns, stay durations, and transition dynamics. For future timestamps, we construct task-specific prompts that encode prediction objectives and temporal context, as detailed in \cref{sec:prompt}. These prompts are processed through frozen pre-trained LLMs to extract high-dimensional semantic representations.

To efficiently integrate these representations, we extract the final hidden state corresponding to the end-of-sequence (\verb|<EOS>|) token, which naturally aggregates contextual information across the entire prompt. Formally, the semantic embedding for segment $i$ is computed as:
\begin{align*}
    \mathbf{TE}_i = \operatorname{SelectLast}(\operatorname{LLM}(\operatorname{Prompt}(x_{(i-1)L+1 :iL}))).
\end{align*}
Task-oriented embeddings follow a similar formulation: $\mathbf{TE}_{\mathcal{T}} = \operatorname{SelectLast}(\operatorname{LLM}(\operatorname{Prompt}(\mathcal{T})))$. Notably, all semantic embeddings are pre-computed offline, transforming what would be computationally intensive online inference into a one-time preprocessing step. This design enables RHYTHM to leverage the semantic understanding of LLMs without incurring runtime computational penalties.

\begin{table*}[!h]
   \centering
   \begin{minipage}{0.24\linewidth}
   \caption{\textbf{Performance evaluation on the Kumamoto, Sapporo, and Hiroshima datasets.}
    We report Accuracy@k across multiple k thresholds (variance $\le2\%$). The highest values are \textbf{bolded}, and the second-best values are \underline{underlined}. RHYTHM consistently outperforms all baselines in the majority of configurations.
    }
   \label{tab:performance}
   \end{minipage}
   \hfill
   \begin{minipage}{0.74\linewidth}
   \resizebox{\textwidth}{!}{%
   \begin{tabular}{l|ccc|ccc|ccc}
       \toprule
       & \multicolumn{3}{c|}{\textbf{Kumamoto}} & \multicolumn{3}{c|}{\textbf{Sapporo}} & \multicolumn{3}{c}{\textbf{Hiroshima}} \\
       \cmidrule(lr){2-4} \cmidrule(lr){5-7} \cmidrule(lr){8-10}
       \textbf{Model} & Acc@1 & Acc@3 & Acc@5 & Acc@1 & Acc@3 & Acc@5 & Acc@1 & Acc@3 & Acc@5 \\
       \midrule
       LSTM & 0.2652 & 0.4799 & 0.5472 & 0.2310 & 0.3940 & 0.4526 & 0.2129 & 0.3775 & 0.4415 \\
       DeepMove & 0.2779 & 0.4986 & 0.5683 & 0.2825 & 0.4672 & 0.5264 &  0.2804 & 0.4810 & 0.5477  \\
       PatchTST & 0.2751 & 0.5018 & 0.5716 & 0.2703 & 0.4582 & 0.5168 & 0.2752 & 0.4839 & 0.5522 \\
       iTransformer & 0.2609 & 0.4724 & 0.5412 & 0.2696 & 0.4500 & 0.5070 & 0.2804 & 0.4857 & 0.5523 \\
       TimeLLM & 0.2712 & 0.4848 & 0.5535 & 0.2792 & 0.4746 & 0.5352 & 0.2698 & 0.4753 & 0.5426 \\
       CMHSA & 0.2862 & 0.5182 & 0.5887 & 0.2890 & \textbf{0.4901} & \underline{0.5525} & 0.2874 & 0.5001 & 0.5684 \\
       PMT & 0.2697 & 0.4475 & 0.5187 & 0.2878 & \underline{0.4896} & 0.5522 & 0.2850 & 0.4982 & 0.5668 \\
       COLA & 0.2864 & 0.5186 & 0.5896 & 0.2847 & 0.4865 & 0.5497 & 0.2874 & 0.5013 & 0.5708 \\
       ST-MoE-BERT & 0.2862 & 0.5155 & 0.5871 & 0.2869 & 0.4856 & 0.5480 & 0.2839 & 0.4925 & 0.5601 \\
       Mobility-LLM & 0.2666 & 0.4793 & 0.5448 & 0.2838 & 0.4703 & 0.5288 & 0.2826 & 0.4856 & 0.5525 \\
       \midrule
       RHYTHM-LLaMA-1B  & \underline{0.2929} & \underline{0.5200} & 0.5835 & 0.2931 & 0.4876 & 0.5502 & 0.2913 & 0.5027 & 0.5753 \\
       RHYTHM-Gemma-2B & 0.2923 & 0.5191 & \underline{0.5932} & \textbf{0.2943} & 0.4896 & \textbf{0.5545} & \textbf{0.2953} & \textbf{0.5074} & \textbf{0.5798} \\
       RHYTHM-LLaMA-3B & \textbf{0.2941} & \textbf{0.5205} & \textbf{0.5947} & \underline{0.2938} & 0.4875 & 0.5523 & \underline{0.2929} & \underline{0.5032} & \underline{0.5756} \\
       \bottomrule
   \end{tabular}%
   }
    \end{minipage}
   \vspace{-0.15in}
\end{table*}

\paragraph{Cross-representational Mobility Prediction.}
Given that the LLM's representation space naturally accommodates both temporal and semantic modalities, we integrate semantic information directly into the temporal representations without increasing the sequence length.
The unified representation $\mathbf{CE}_i$ for segment $i$ is constructed by element-wise summation of the segment representation $\tilde{\mathbf{SE}}_i$ and its corresponding semantic encoding $\mathbf{TE}_i$:
\begin{align*}
    \mathbf{CE}_i = \tilde{\mathbf{SE}}_i + \mathbf{TE}_i.
\end{align*}
For future time points, we combine temporal and task-specific representations: $\mathbf{CE}_{N+j} = \tilde{\mathbf{E}}_{T+j} + \mathbf{TE}_{\mathcal{T}}$.

These combined representations $\mathbf{CE}_i$ are subsequently processed by a frozen pre-trained LLM.
The LLM transforms these inputs through its multi-layer architecture, leveraging its pre-trained knowledge to perform contextual reasoning over the fused spatio-temporal and semantic signals, producing final layer hidden states $h_i$:
\begin{align*}
    h_i = \operatorname{LLM}(\mathbf{CE}_i).
\end{align*}

A learnable projection head then transforms these representations into location-specific scores:
\begin{align*}
    P(l_{T+j}|\mathcal{X},\mathcal{T}) = \operatorname{softmax}(W_o\tilde{\mathbf{h}}_{N+j} + \mathbf{b}_o),
\end{align*}
where $W_o \in \mathbb{R}^{|\mathcal{L}|\times D}$ maps from hidden dimension to location vocabulary.
This probability distribution on candidate locations enables the model to generate mobility predictions.

\vspace{-.5em}
\subsection{Computational Efficiency}
RHYTHM's architectural design incorporates multiple strategies to optimize both computational efficiency and parameter utilization. Pre-computation of semantic representations using the frozen LLM occurs as a one-time preprocessing step, completely removing language model inference from the training and inference pipeline. Concurrently, the temporal segmentation strategy compresses the operational sequence length from $T+H$ to $N+H$, yielding a reduction in attention complexity from $\mathcal{O}((T+H)^2)$ to $\mathcal{O}((N+H)^2)$—a critical optimization for handling lengthy mobility traces.
Additionally, by maintaining frozen LLM parameters throughout training, we achieve accelerated convergence while minimizing memory footprint. This confluence of design decisions empowers RHYTHM to handle extensive trajectory sequences without compromising prediction accuracy (demonstrated in~\cref{fig:efficiency}), rendering it particularly well-suited for deployment in resource-constrained environments and large-scale mobility applications.

\vspace{-1em}
\section{Experiment}
\label{sec:exp}

\paragraph{Models.}
We assess RHYTHM's mobility prediction capabilities using various pre-trained LLMs as backbones, sourced from Hugging Face with their original weights. Specific LLM configurations are listed in~\cref{ap:llm_variant}.

\paragraph{Evaluation Metrics.}
We measure ranking quality using Accuracy@k and Mean Reciprocal Rank (MRR), complemented by Dynamic Time Warping (DTW) \cite{muller2007dynamic} and BLEU \cite{papineni2002bleu} for trajectory-level evaluation. Metric definitions are provided in~\cref{subsection:metrics}.

\paragraph{Datasets.}
Our experiments utilize three urban mobility datasets from Kumamoto, Sapporo, and Hiroshima, obtained from YJMob100K~\cite{yabe2024yjmob100k}. 
Days are discretized into 48 half-hour intervals, with sparse observations across time slots.
We partition data chronologically into training (70\%), validation (20\%), and test (10\%) splits. Dataset specifications are detailed in~\cref{section:dataset}.

\paragraph{Baselines.}
We benchmark RHYTHM against established baselines, including LSTM~\cite{kong2018hst}, DeepMove~\cite{feng2018deepmove}, PatchTST~\cite{nie2023a}, iTransformer~\cite{liu2024itransformer}, TimeLLM~\cite{jin2024timellm}, PMT~\cite{wu2024pretrained}, ST-MoE-BERT~\cite{he2024st}, CMHSA~\cite{hong2023context}, COLA~\cite{wang2024cola}, and Mobility-LLM~\cite{gong2024mobility}. See~\cref{subsection:baselines} for baseline details.

\paragraph{Results.}
\cref{tab:performance} demonstrates that RHYTHM consistently surpasses baseline methods on most criteria across all cities.
For Sapporo and Hiroshima, RHYTHM attains superior performance across all metrics.
These results highlight RHYTHM's robust capability for mobility forecasting.
The competitive performance of CMHSA and PMT in Accuracy@3 for Kumamoto can be attributed to their tailored attention architectures, which are particularly adept at identifying intermediate-ranked location candidates within this specific geographical area.
Although Mobility-LLM employs an LLM-based framework, its performance falls short of RHYTHM, primarily due to its original optimization for visit intention prediction that emphasizes semantic understanding.
By comparison, RHYTHM combines temporal segmentation with LLM capabilities to capture hierarchical spatio-temporal patterns, emphasizing accurate location probability estimation.
This architectural emphasis positions RHYTHM to achieve superior performance in ranking metrics.
In summary, RHYTHM delivers a 2.4\% gain in Accuracy@1 and 1.0\% improvement in Accuracy@5 relative to the strongest competing approach.

Further empirical analyses including spatial accuracy assessments, temporal pattern evaluation across daily and weekly cycles, computational efficiency benchmarks, model scaling experiments, and component-wise ablation experiments are presented in~\cref{sec:extend_exp} given space limitations.

\vspace{-1em}
\section{Conclusion}
\label{sec:conclusion}
We present RHYTHM, a computationally efficient architecture for  mobility prediction that employs temporal segmentation to encode spatio-temporal relationships and utilizes semantic representations to model periodic behaviors. Through the incorporation of frozen pre-trained LLMs as contextual reasoning modules, RHYTHM captures the underlying decision dynamics—especially for non-routine trajectories—while maintaining computational tractability. Furthermore, the framework's modular design facilitates straightforward adaptation across different pre-trained language models without architectural modifications.

\newpage
\section*{Broader Impact}
This paper presents a novel foundation model architecture for human mobility analysis, seeking to enhance the robustness and transferability of foundation models within spatio-temporal contexts. Although direct societal impacts are not immediately apparent, this work establishes fundamental capabilities for subsequent developments in urban infrastructure design, epidemiological monitoring, and transportation systems. Nevertheless, the model carries the risk of perpetuating or intensifying inherent biases within the training datasets, which could result in disparate prediction quality across different demographic groups or geographic regions.

\label{ap:broader}

\section*{Acknowledgments}
The authors would like to thank the anonymous reviewers and program chairs for constructive comments.

H.H. and Q.R.W.'s work is supported by the National Science Foundation (NSF) under Grant Nos. 2125326, 2114197, 2228533, and 2402438, as well as by the Northeastern University iSUPER Impact Engine. 
H.L. is partially supported by the OpenAI Researcher Access Program. 
This research was supported in part through the computational resources and staff contributions provided for the Quest high performance computing facility at Northwestern University which is jointly supported by the Office of the Provost, the Office for Research, and Northwestern University Information Technology.
Any opinions, findings, conclusions, or recommendations expressed in the paper are those of the author and do not necessarily reflect the views of the funding agencies.

\bibliography{main}
\bibliographystyle{icml2025}

\newpage  %

\clearpage

\newpage
\normalsize

\onecolumn
\appendix
\part*{Supplementary Material}
\label{sec:append}
{
\setlength{\parskip}{-0em}
\startcontents[sections]
\printcontents[sections]{ }{1}{}
}

\section{Limitations}
\label{sub:limitation}

Several important limitations warrant consideration regarding RHYTHM.
First, the model's effectiveness depends significantly on the underlying pretrained LLMs, which were initially developed for natural language processing rather than spatial-temporal modeling.
When pretrained models suffer from resource constraints, their ability to effectively encode human movement patterns may be compromised.
Moreover, RHYTHM currently employs a non-autoregressive prediction paradigm for mobility forecasting.
Although recent research demonstrate promising results with autoregressive formulations for temporal sequences \cite{liu2024autotimes}, incorporating these techniques into RHYTHM represents a valuable direction for future development.
Additionally, despite utilizing frozen LLMs to enhance computational efficiency, RHYTHM's training duration remains substantial, potentially limiting its deployment in time-sensitive scenarios.
Nevertheless, RHYTHM establishes an innovative foundation model architecture specifically designed for mobility prediction, demonstrating improvements in both computational efficiency and forecast accuracy.
This work provides a foundation for subsequent advances in model scalability and resource optimization.
Future efforts will focus on developing sophisticated fine-tuning strategies to enhance performance while minimizing computational requirements, leveraging recent advances in parameter-efficient adaptation and model compression techniques \cite{luo2025fast,hu2024outlier,dettmers2024qlora,xiao2023smoothquant}.

\section{Related Work}
\label{sec:background}
\paragraph{Mobility Prediction.}
The field of human mobility prediction has progressed from classical statistical approaches to sophisticated deep learning architectures.
Mechanistic models including the gravity framework~\cite{cabanas2025human} and radiation theory~\cite{simini2012universal} forecast collective movement patterns through distance-decay and intervening opportunities, yet cannot capture individual-specific behaviors.
Addressing this limitation, stochastic methods such as Markovian models~\cite{gambs2012next}, decision tree techniques~\cite{he2023forecasting}, and matrix decomposition~\cite{yang2014modeling} have been developed to predict user-specific location sequences.
Despite enabling personalized predictions, these approaches face challenges with data sparsity and complex temporal interdependencies characteristic of mobility data.
Neural architectures introduced sequential modeling through LSTMs \cite{liu2016predicting}, enabling temporal context understanding, with attention-augmented extensions~\cite{feng2018deepmove} mitigating gradient degradation. Nonetheless, such models frequently miss periodic behavioral patterns.
Integrated frameworks including Graph-Flashback~\cite{rao2022graph} and GCDAN~\cite{dang2022predicting} incorporate network topology for spatial modeling, though fixed-window constraints hinder their long-horizon prediction capabilities.
The Transformer paradigm~\cite{vaswani2017attention} transformed sequence modeling through self-attention for capturing distant dependencies.
Notable extensions include STAN~\cite{luo2021stan}, which integrates spatial-temporal attention for location recommendation, COLA~\cite{wang2024cola}, which generalizes across urban environments, and GETNext~\cite{yang2022getnext}, which separates personal patterns from collective trends.
However, these Transformer variants maintain timestamp-based representations, resulting in quadratic computational growth for extended trajectories and lacking explicit modeling of nested temporal cycles (daily within weekly patterns).
Contemporary research investigates large language models (LLMs) for mobility applications, exploiting their robust transfer learning properties \cite{gong2024mobility, liu2024human}. Implementations such as LLM-Mob~\cite{wang2023would} and AgentMove~\cite{feng2024agentmove} utilize strategic prompting for location forecasting and user identification, while TrajGPT~\cite{hsu2024trajgpt} employs generative modeling for trajectory synthesis. Yet these methods process mobility data as undifferentiated token sequences, disregarding inherent temporal structures and the representational disconnect between linguistic and spatio-temporal modalities. In contrast to current LLM approaches that handle trajectories as generic sequences, our method implements temporal segmentation to directly encode periodic structures (daily/weekly patterns), addressing representational gaps and modeling hierarchical temporal relationships for enhanced long-range mobility forecasting.

\paragraph{Time Series Foundation Models.}
Current time series foundation models fall into two primary paradigms: transformer-centric architectures and language model adaptations.
Transformer-centric approaches \cite{wu2024stanhop, liu2024itransformer,nie2023a} emphasize architectural innovations and attention mechanisms for temporal pattern extraction.
Notably, PatchTST \cite{nie2023a} employs segment-wise attention to model extended temporal relationships, while STanHop \cite{wu2024stanhop} and Crossformer \cite{zhang2023crossformer} utilize multi-level attention architectures to encode both sequential patterns and hierarchical temporal organization.
Language model adaptations \citep{liu2024autotimes,jin2024timellm} repurpose pre-trained LLMs for time series analysis, demonstrating competitive performance in forecasting benchmarks.
Specifically, AutoTime \citep{liu2024autotimes} develops an autoregressive framework tailored for sequential dependency modeling, while TimeLLM \cite{jin2024timellm} harnesses LLMs to learn complex temporal state transitions.
Nevertheless, these approaches cannot adequately handle the unique characteristics of human mobility—particularly sudden spatial transitions and irregular temporal patterns. In contrast, \sys combines specialized spatio-temporal representations with hierarchical modeling to capture these complex mobility dynamics effectively.

\paragraph{Cross-domain Adaptation of LLMs.}
LLMs transform from domain-specific language processors into general-purpose foundation models with advanced reasoning abilities spanning multiple disciplines \cite{alabdulmohsin2022revisiting,brown2020language}. The combination of transformer architectures and large-scale pretraining enables exceptional knowledge transfer beyond textual domains. For visual understanding, frameworks like CLIP \cite{radford2021learning} create unified vision-language representations enabling zero-shot classification, while temporal modeling approaches including One-Fits-All \cite{zhou2023one} and LLM4TS \cite{chang2025llm4ts} achieve strong forecasting results by representing numerical data as discrete tokens. Within biomedical applications, specialized models such as BioBERT \cite{lee2020biobert} and BioGPT \cite{luo2022biogpt} show substantial improvements on medical text processing, with instruction-aligned systems like Med-PaLM reaching near-expert performance on clinical questions \cite{singhal2023large}. Financial domain adaptations including FinBERT \cite{huang2023finbert} and BloombergGPT \cite{wu2023bloomberggpt} demonstrate marked advantages over generic models for market sentiment and entity extraction tasks.

To avoid resource-intensive complete retraining, efficient adaptation methods have emerged as preferred strategies. Low-Rank Adaptation (LoRA) \cite{hu2021loralowrankadaptationlarge} augments attention mechanisms with decomposed weight updates, while alternative strategies maintain frozen model parameters through modality-specific input transformations. Visual adaptation techniques \cite{alayrac2022flamingo,tsimpoukelli2021multimodal} learn compact encoders that generate conditioning signals for static LLMs, whereas temporal sequence methods \cite{liu2024autotimes,jin2024timellm} utilize learned projections for numerical-to-embedding conversion.

LLM applications in mobility analysis remain nascent, with current methods requiring substantial parameter updates. Mobility-LLM \cite{gong2024mobility} updates model subsets during training, while LLM-Mob \cite{wang2023would} employs contextual prompting without explicit temporal structure. Conversely, RHYTHM preserves all LLM parameters unchanged, retaining pretrained capabilities while implementing a custom spatio-temporal encoding architecture optimized for mobility sequence processing.

\section{Attention Implementation Details}\label{section:attention}
We employ a pre-normalized transformer design that promotes training stability, augmented with gated feed-forward modules for enhanced representational capacity. The architectural formulation follows:
\begin{align*}
Z &= \text{LayerNorm}(X) + \text{Multi-Head Attention}(\text{LayerNorm}(X)), \\
\tilde{Z} &= Z + \text{GatedFFN}(\text{LayerNorm}(Z)),
\end{align*}
with $X$ denoting the input representations. The multi-head attention mechanism is defined as:
\begin{align*}
\text{Multi-Head}(X) &= [\text{head}_1 \| \text{head}_2 \| \ldots \| \text{head}_h] W_{\text{out}}, \\
\text{head}_i &= \text{Softmax}\left(\frac{X W_{q,i} (X W_{k,i})^\top}{\sqrt{d_k}}\right) X W_{v,i},
\end{align*}
where we utilize $h$ parallel attention heads, with $W_{q,i}, W_{k,i}, W_{v,i} \in \mathbb{R}^{d \times d_k}$ representing query, key, and value transformations for head $i$, and $W_{\text{out}} \in \mathbb{R}^{d \times d}$ serving as the final projection. The gated feed-forward component employs dynamic modulation:
\begin{align*}
\text{GatedFFN}(Z) &= \text{FFN}(Z) \odot \sigma(W_{\text{gate}} Z), \\
\text{FFN}(Z) &= W_2 \, \text{GELU}(W_1 Z),
\end{align*}
where $\sigma$ represents sigmoid activation, $\odot$ indicates Hadamard product, and $W_{\text{gate}} \in \mathbb{R}^{d \times d}$ controls the gating behavior. The feed-forward transformation implements a 4× dimension expansion through $W_1 \in \mathbb{R}^{4d \times d}$ and compression via $W_2 \in \mathbb{R}^{d \times 4d}$. Regularization through dropout follows both attention and feed-forward computations to mitigate overfitting.

\section{Theoretical Guarantee}
\label{sub:theory}
Our architectural decisions are grounded in robust theoretical foundations.
Utilizing an LLM for universal sequence representation extraction offers two theoretical benefits: (1) convergence guarantees for model outputs, established by \citet[Theorem~E.2]{zhou2023one}, and (2) uniform feature distribution properties within the LLM's final hidden representations, proven in \citet[Theorem~E.3]{zhou2023one}.
These characteristics collectively strengthen the downstream MLP classifier's learning capacity.
Furthermore, given our transformer-based architecture, \citet{ramsauer2020hopfield} establishes that transformers constitute a specific instantiation of contemporary Hopfield networks.
This connection provides bounded memory retrieval guarantees for our LLM-based approach, as formalized in \citet[Lemma~3.2]{hu2024computational}.
Such theoretical underpinnings substantiate our design choices, with empirical results confirming these theoretical predictions.

\section{Example Prompt}\label{sec:prompt}

\begin{tcolorbox}[title=Trajectory Information]
\footnotesize\ttfamily
This is the trajectory of user <User\_ID> of day <Day\_ID> which is a <Day\_of\_Week>. The trajectory consists of <N> records, each record of coordinate is as follows: 08:30: (X=136, Y=42); 09:00: (X=136, Y=42); 09:30: (X=137, Y=41); 10:00: (X=146, Y=37); 10:30: (X=145, Y=38); 11:00: (X=144, Y=38); 11:30: (X=135, Y=41); 12:00: (X=135, Y=42); 12:30: (X=135, Y=42); 13:00: (X=135, Y=42).

\medskip

Key transitions: At 10:00: (X=137, Y=41) → (X=146, Y=37); At 11:30: (X=144, Y=38) → (X=135, Y=41).

\medskip

Main stay locations: (X=136, Y=42) from 08:30 to 09:30 (0.5 hours); (X=145, Y=38) from 10:00 to 11:00 (0.5 hours); (X=135, Y=42) from 11:30 to 13:00 (1.5 hours).
\end{tcolorbox}

\begin{tcolorbox}[title=Task Description]
\footnotesize\ttfamily
You are a mobility prediction assistant that forecasts human movement patterns in urban environments. The city is represented as a 200 x 200 grid of cells, where each cell is identified by coordinates (X,Y). The X coordinate increases from left (0) to right (199), and the Y coordinate increases from top (0) to bottom (199).

\medskip

TASK: Based on User <User\_ID>'s historical movement patterns, predict their locations for Day <Day\_ID> (<Day\_of\_Week>). The predictions should capture expected locations at 30-minute intervals throughout the day (48 time slots). The model should analyze patterns like frequent locations, typical daily routines, and time-dependent behaviors to generate accurate predictions of where this user is likely to be throughout the next day.

\medskip

The previous days' trajectory data contains information about the user's typical movement patterns, regular visited locations, transition times, and duration of stays. Key patterns to consider include: home and work locations, morning and evening routines, lunch-time behaviors, weekend vs. weekday differences, and recurring visit patterns.
\end{tcolorbox}

\section{Dataset}\label{section:dataset}
\cref{tab:dataset} presents comprehensive statistics for the three datasets utilized in our experiments.

\begin{table}[H]
    \centering
    \caption{Dataset Statistics}
    \label{tab:dataset}
    \begin{tabular}{lcccc}
        \toprule
        \textbf{City} & \textbf{Users} & \textbf{Duration} & \textbf{Spatial Resolution} & \textbf{Places}  \\
        \midrule
        Kumamoto  & 3k   & 75 days      & 500m $\times$ 500m        & 40k\\
        Sapporo   & 17k  & 75 days  & 500m $\times$ 500m & 40k \\
        Hiroshima    & 22k  & 75 days  & 500m $\times$ 500m     & 40k   \\
        \bottomrule
    \end{tabular}
\end{table}

\section{Experiment Setting}

\subsection{Evaluation Metrics}\label{subsection:metrics}
\textbf{Accuracy@k} quantifies the fraction of instances where true locations appear among the top-$k$ predictions:
\begin{align*}
   \text{Accuracy@}k = \frac{1}{H}\sum_{i=1}^{H} \mathbbm{1}(l_{T+i} \in \text{top-}k(\hat{p}_{T+i})),
\end{align*}
with $\mathbbm{1}(\cdot)$ denoting the indicator function and $\hat{p}_{T+i}$ representing the output probability vector.

\textbf{Mean Reciprocal Rank (MRR)} assesses ranking quality through reciprocal positions:
\begin{align*}
   \text{MRR} = \frac{1}{H}\sum_{i=1}^{H} \frac{1}{\text{rank}(l_{T+i})},
\end{align*}
where $\text{rank}(l_{T+i})$ indicates the ordinal position of the actual location.

\textbf{Dynamic Time Warping (DTW)} computes trajectory alignment cost:
\begin{align*}
   \text{DTW}(\mathcal{Y}, \hat{\mathcal{Y}}) = \min_{\pi} \sum_{(i,j) \in \pi} d(l_{T+i}, \hat{l}_{T+j}),
\end{align*}
with $\pi$ defining an optimal alignment path and $d(\cdot,\cdot)$ measuring Euclidean separation.

\textbf{BLEU} evaluates sequence similarity via n-gram correspondences:
\begin{align*}
   \text{BLEU} = BP \cdot \exp\left(\sum_{n=1}^{N} w_n \log p_n\right),
\end{align*}
where $p_n$ captures n-gram matching rates, $w_n$ assigns importance weights, and $BP$ adjusts for length discrepancies.

\subsection{Settings}
Our experiments employ 30-minute temporal granularity. We configure a historical context of 7 days comprising 336 time intervals, with predictions extending over 48 intervals (equivalent to 24 hours).
The segment size is fixed at 48 time intervals throughout all experiments.

\subsection{Baselines}\label{subsection:baselines}
We benchmark RHYTHM against three categories of baseline methods: LSTM-based architectures, transformer-based approaches, and LLM-based frameworks.
Among transformer architectures, we evaluate PatchTST~\cite{nie2023a}, PMT~\cite{wu2024pretrained}, ST-MoE-BERT~\cite{he2024st}, CMHSA~\cite{hong2023context}, iTransformer~\cite{liu2024itransformer}, and COLA~\cite{wang2024cola}.
Within this group, ST-MoE-BERT, PMT, and COLA represent current best-performing methods in mobility forecasting, while PatchTST and iTransformer are leading general-purpose temporal sequence models.
To ensure equitable evaluation, we augment these general time series models with spatio-temporal encodings.
For LLM-based comparisons, we include TimeLLM~\cite{jin2024timellm} and Mobility-LLM~\cite{gong2024mobility}.
TimeLLM achieves state-of-the-art results in LLM-driven time series prediction, which we enhance with spatio-temporal representations for consistent comparison.
Mobility-LLM provides a comprehensive LLM framework supporting diverse mobility-related tasks.
Our LSTM baselines comprise the classical LSTM~\cite{kong2018hst} and DeepMove~\cite{feng2018deepmove} architectures.

\subsection{Computational Resource}
\label{ap:resource}
All experiments are conducted on a single NVIDIA A100 GPU equipped with 40GB memory, paired with a 24-core Intel(R) Xeon(R) Gold 6338 processor running at 2.00GHz.
Our implementation leverages PyTorch~\cite{paszke2019pytorch} and integrates the Hugging Face Transformers library for model deployment.

\subsection{Hyperparameters}
We detail the training configurations employed across all models.
The embedding dimensions are configured as follows: time-of-day and day-of-week embeddings utilize 128 dimensions each, categorical location representations use 256 dimensions, and coordinate projections employ 128 dimensions.
Training optimization is performed using \textbf{AdamW}~\cite{loshchilov2017decoupled}.
We perform comprehensive hyperparameter optimization by evaluating learning rates within $\{1 \times 10^{-4}, 3 \times 10^{-4}, 5 \times 10^{-4}\}$ and weight decay parameters from $\{0, 0.001, 0.01\}$.
Optimal hyperparameters for each dataset are identified through rigorous validation experiments. To ensure fair evaluation, all models employ a uniform batch size of 64. Final parameter selections are determined by validation set performance.

\subsection{LLM variants}\label{ap:llm_variant}
We incorporated various pre-trained language models as text embedders and fixed backbone architectures in RHYTHM to assess performance across different model scales. \cref{tab:llm_variants} lists the foundation models utilized via the Hugging Face Transformers library, spanning parameter counts from 125M to 3B.
\begin{table}[h]
\centering
\caption{List of LLMs used in RHYTHM.}
\label{tab:llm_variants}
\begin{tabular}{lrl}
\toprule
\textbf{Model} & \textbf{Parameters} & \textbf{HuggingFace Repository} \\
\midrule
OPT-125M & 125M & \href{https://huggingface.co/facebook/opt-125m}{facebook/opt-125m} \\
OPT-350M & 350M & \href{https://huggingface.co/facebook/opt-350m}{facebook/opt-350m} \\
LLaMA-3.2-1B & 1.24B & \href{https://huggingface.co/meta-llama/Llama-3.2-1B}{meta-llama/Llama-3.2-1B} \\
Qwen-2.5-1.5B & 1.54B & \href{https://huggingface.co/Qwen/Qwen2.5-1.5B}{Qwen/Qwen2.5-1.5B} \\
DeepSeek-R1-1.5B & 1.78B & \href{https://huggingface.co/deepseek-ai/DeepSeek-R1-Distill-Qwen-1.5B}{deepseek-ai/DeepSeek-R1-Distill-Qwen-1.5B} \\
Gemma-2-2B & 2.61B & \href{https://huggingface.co/google/gemma-2-2b-it}{google/gemma-2-2b-it} \\
Phi-2 & 2.78B & \href{https://huggingface.co/microsoft/phi-2}{microsoft/phi-2} \\
LLaMA-3.2-3B & 3.21B & \href{https://huggingface.co/meta-llama/Llama-3.2-3B}{meta-llama/Llama-3.2-3B} \\
\bottomrule
\end{tabular}%
\end{table}

\section{Extended Experiments}\label{sec:extend_exp}

\subsection{Geographical Evaluation}
We assess RHYTHM with baselines using BLEU and DTW metrics, measuring sequence correspondence and trajectory alignment accuracy, respectively.
\cref{tab:bleu_dtw} reveals that RHYTHM achieves optimal DTW scores on Sapporo, indicating superior spatial trajectory matching.
Although COLA attains highest BLEU values across all datasets, RHYTHM secures second place for Kumamoto.
These results reveal an inherent balance between precise sequence replication and spatial accuracy optimization.
This performance difference may stem from COLA's posterior correction mechanism, which adjusts predictions toward observed location frequency distributions, potentially improving intermediate-rank predictions by reducing bias toward high-frequency locations.
RHYTHM substantially surpasses LSTM architectures and transformer models through its temporal segmentation and semantic embedding strategies, achieving improved sequence consistency and location precision.
This combination yields well-balanced performance for practical mobility applications.
Regarding MRR evaluation, RHYTHM uniformly exceeds all competing methods with a 1.44\% gain over the strongest alternative, confirming its robust ranking performance across varied movement patterns.

\begin{table*}[ht]
   \centering
   \caption{\textbf{Geographical assessment of RHYTHM against baseline methods.} We report DTW ($\downarrow$), BLEU ($\uparrow$), and MRR ($\uparrow$) metrics with variance below $2\%$. Best performance is shown in \textbf{bold}, with second-best results \underline{underlined}.}
   \label{tab:bleu_dtw}
   \resizebox{.8\textwidth}{!}{%
   \begin{tabular}{l|ccc|ccc|ccc}
       \toprule
       & \multicolumn{3}{c|}{\textbf{Kumamoto}} & \multicolumn{3}{c|}{\textbf{Sapporo}} & \multicolumn{3}{c}{\textbf{Hiroshima}} \\
       \cmidrule(lr){2-4} \cmidrule(lr){5-7} \cmidrule(lr){8-10}
       \textbf{Model} & DTW & BLEU & MRR & DTW & BLEU & MRR & DTW & BLEU & MRR \\
       \midrule
       LSTM & 5014 & 0.1564 & 0.3860 & 4507 & 0.1716 & 0.3270 & 5908 & 0.1544 & 0.3113 \\
       DeepMove & 4630 & 0.1746 & 0.4021 & 3818 & 0.1959 & 0.3887 & 4981  & 0.1933 & 0.3959 \\
       PatchTST & 5251 & 0.1315 & 0.4021 & 4099 & 0.1784 & 0.3773 & 5021 & 0.1884 & 0.3945 \\
       iTransformer & 6178 & 0.1275 & 0.3796 & 4074 & 0.1780 & 0.3730 & 5094 & 0.1789 & 0.3977 \\
       TimeLLM & 5984 & 0.1285 & 0.3912 & 3915 & 0.2145 & 0.3902 & 5126 & 0.1988 & 0.3872  \\
       CMHSA & 4490 & 0.1810 & 0.4158 & \underline{3786} & 0.2299 & 0.4034 & \underline{4841} & \underline{0.2289} & 0.4086 \\
       PMT & 4536 & 0.1524 & 0.3720 & 3799 & 0.2017 & 0.4026 & 4851 & 0.2009 & 0.4065 \\
       COLA & \underline{4446} & \textbf{0.2064} & 0.4164 & 3793 & \textbf{0.2496} & 0.3996 & \textbf{4840} & \textbf{0.2445} & 0.4095 \\
       ST-MoE-BERT & 4691 & 0.1557 & 0.4151 & 3796 & 0.2102 & 0.4001 & 4889 & 0.2117 & 0.4031 \\
       Mobility-LLM & 5603 & 0.1649 & 0.3858 & 3911 & 0.1917 & 0.3902 & 4985 & 0.2056 & 0.3990 \\
       \midrule
       RHYTHM-LLaMA-1B & 4478 & 0.1793 & \underline{0.4216} & \textbf{3745} & \textbf{0.2496} & 0.4045 & 5059 & 0.2083 & 0.4069 \\
       RHYTHM-Gemma-2B & \textbf{4416} & \underline{0.1928} & 0.4205 & 3995 & 0.2019 & \textbf{0.4065} & 4857 & 0.2109 & \textbf{0.4173} \\
       RHYTHM-LLaMA-3B & 4470 & 0.1814 & \textbf{0.4220} & 4035 & 0.1917 & \underline{0.4048} & 4935 & 0.2093 & \underline{0.4140} \\
       \bottomrule
   \end{tabular}%
   }
\end{table*}

\subsection{Daily and Weekly Trend Analysis}
\begin{figure}[ht]
    \centering
    \includegraphics[width=0.99\linewidth]{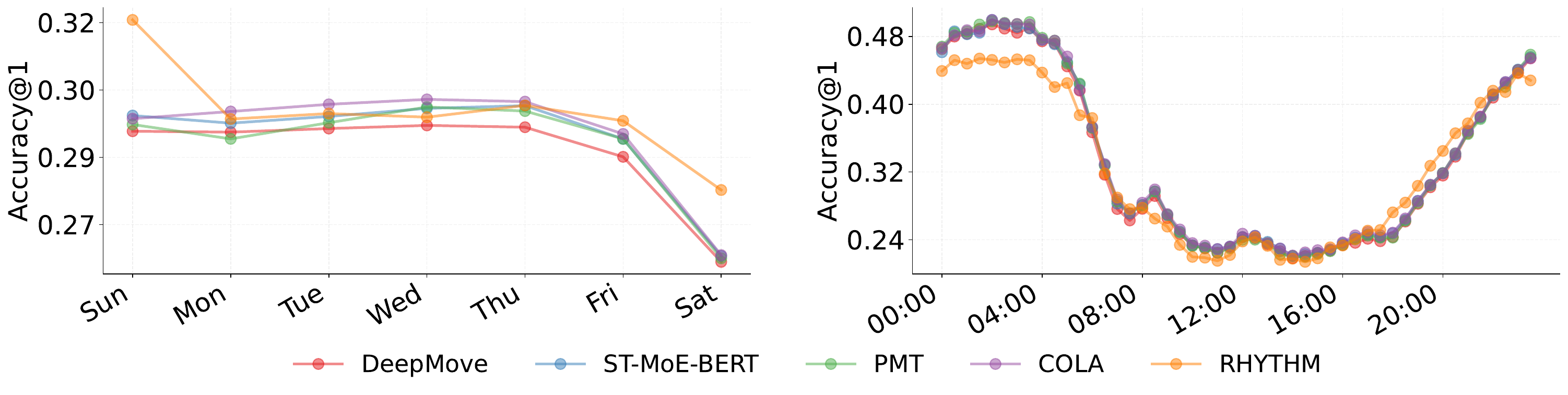}
    \caption{\textbf{Temporal performance patterns of RHYTHM and baselines on Sapporo data showing weekly (left) and daily (right) variations.} The results demonstrate systematic performance fluctuations across both diurnal and weekly cycles.}
    \label{fig:trend}
\end{figure}

We examine temporal performance variations of RHYTHM and competing methods on the Sapporo dataset, analyzing accuracy patterns across daily and weekly cycles (\cref{fig:trend}).
RHYTHM generally surpasses baseline approaches across temporal dimensions, particularly during evening commute periods and weekends, with performance gains of 3.4\% and 5.0\%, respectively. These observations align with~\citet{barbosa2018human}, who documented increased variability and irregular patterns in weekend mobility behaviors. Notably, RHYTHM's advantages diminish during highly predictable movement periods, including overnight hours and typical weekday routines, while demonstrating substantial improvements during less structured timeframes such as weekends and evening transitions when mobility patterns become more stochastic.
This enhanced performance during irregular periods stems from RHYTHM's LLM-powered reasoning mechanisms, which effectively model complex behavioral factors underlying mobility decisions during non-routine scenarios.
Conversely, baseline methods depend on rigid temporal heuristics, constraining their adaptability to fluctuating movement dynamics.

\subsection{Transferability}
To validate RHYTHM's generalizability across different pretrained architectures, we experiment with various backbone model scales and evaluate their mobility prediction performance (detailed in \cref{tab:performance}).
We specifically investigate RHYTHM's behavior when equipped with LLaMA-3.2-1B, LLaMA-3.2-3B, and Gemma-2-2B as backbone models.
Our findings reveal consistent performance improvements with increasing model capacity.
Both LLaMA-3.2-3B and Gemma-2-2B configurations surpass LLaMA-3.2-1B across most evaluation metrics.
These results confirm that RHYTHM's effectiveness scales proportionally with backbone model size, suggesting potential for further gains with larger architectures on expanded datasets.
All models undergo 30 training epochs in our experiments.
The larger LLaMA-3.2-3B variant likely benefits from extended training to reach optimal performance relative to the more efficient LLaMA-3.2-1B.
Despite this constraint, LLaMA-3.2-3B maintains strong competitive performance against its smaller counterpart.
Specifically, LLaMA-3.2-3B yields a 0.40\% Acc@1 improvement over LLaMA-3.2-1B, demonstrating RHYTHM's effective scaling properties.

\begin{figure}[ht] %
    \centering
    \begin{minipage}[b]{0.49\linewidth}
        \centering
        \includegraphics[width=\linewidth]{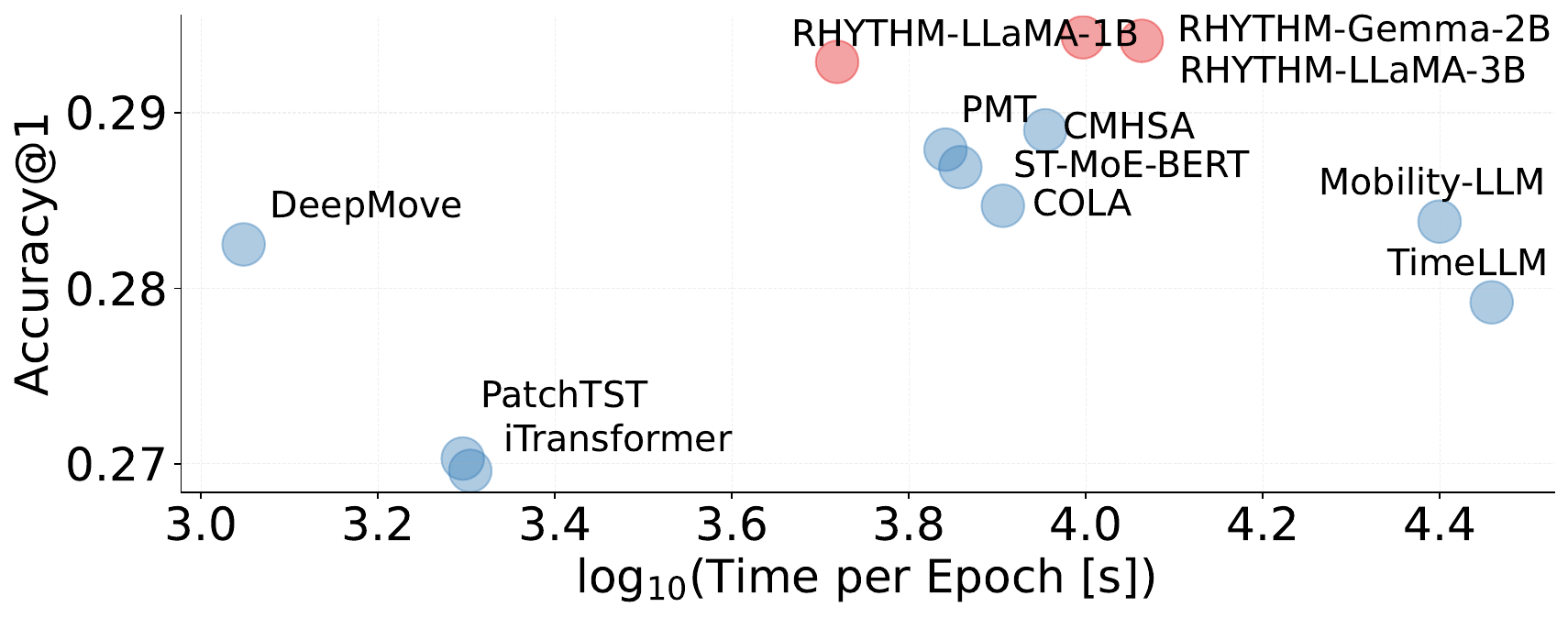} \caption{Computational efficiency versus predictive accuracy trade-offs for RHYTHM and baseline approaches on the Sapporo dataset.}
        \label{fig:efficiency}
    \end{minipage}
    \hfill 
    \begin{minipage}[b]{0.49\linewidth}
        \centering
        \includegraphics[width=\linewidth]{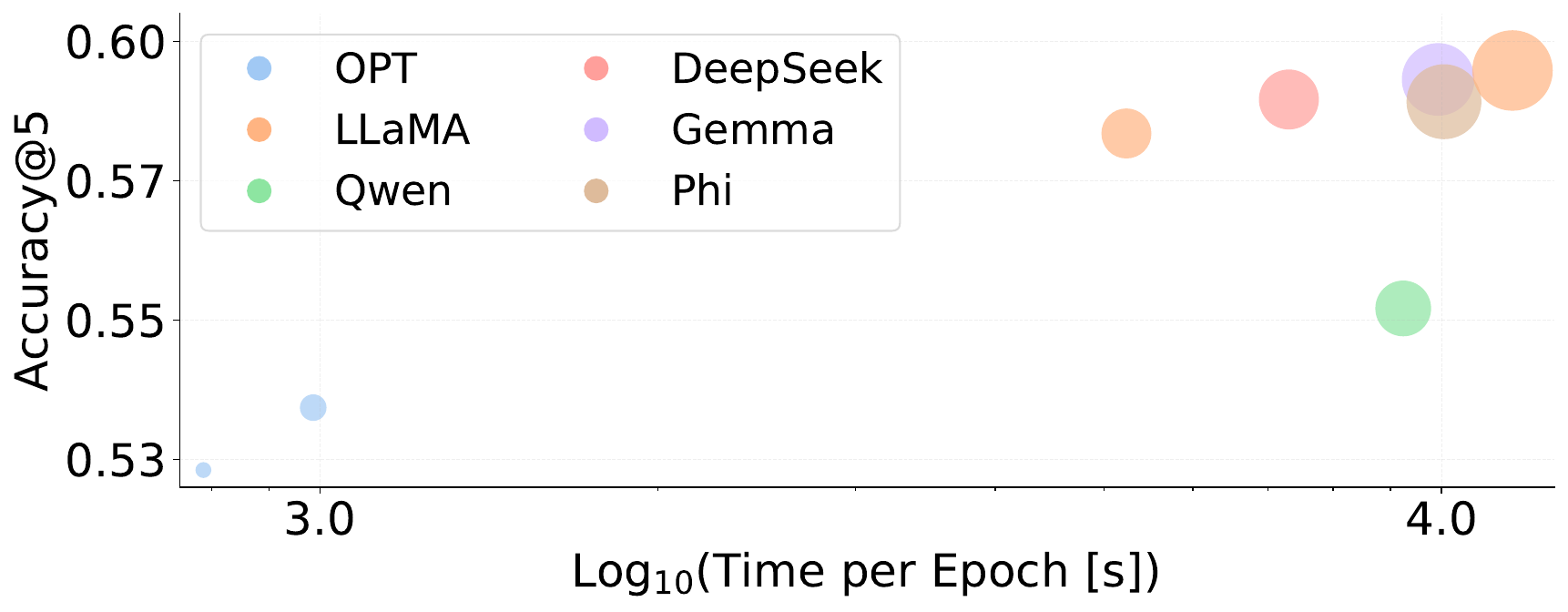}
        \caption{Computational performance across different LLM backbones using identical experimental settings from \cref{tab:scale}.}
        \label{fig:scale}
    \end{minipage}
\end{figure}

\subsection{Training Speed}
We benchmark RHYTHM's computational efficiency on the Sapporo dataset under consistent training settings.
Experiments utilize a single NVIDIA A100 GPU with 40GB memory capacity.
\cref{fig:efficiency} presents the comparative results.
RHYTHM demonstrates superior training efficiency relative to most baseline approaches.
While LSTM, DeepMove, PatchTST, and iTransformer exhibit faster per-epoch training, RHYTHM's accuracy gains justify the modest computational overhead.
Remarkably, RHYTHM achieves training speeds on par with those of PMT, COLA, and ST-MoE-BERT while managing substantially larger parameter budgets, validating its efficient architectural design.
Training time exhibits predictable scaling characteristics: the LLaMA-3B configuration requires 2.2× the computation of LLaMA-1B, while Gemma-2-2B increases the training time by 1.9×.

\subsection{Scaling Behavior}

\begin{table}[t]
    \centering
    \caption{\textbf{Scaling analysis for RHYTHM.} We assess RHYTHM's scaling behavior across pre-trained models with different parameter counts. Evaluation encompasses Accuracy@k, MRR, and per-epoch training duration. Best performance is shown in \textbf{bold}, with second-best results \underline{underlined}. Performance generally scales positively with increased model capacity across configurations.}
    \vspace{0.1in}
    \label{tab:scale}
       \begin{tabular}{lccccc}
        \toprule
        Backbone & Time(s) & Acc@1 & Acc@3 & Acc@5 & MRR \\
        \midrule
        OPT-125M & 787 & 0.2798 & 0.4726 & 0.5231 & 0.3819 \\
        OPT-350M & 986 & 0.2837 & 0.4789 & 0.5343 & 0.3923 \\
        LLaMA-3.2-1B & 5235 & \underline{0.2929} & \underline{0.5200} & 0.5835 & \underline{0.4216} \\
        Qwen-2.5-1.5B & 9241 & 0.2897 & 0.4873 & 0.5521 & 0.4049 \\
        DeepSeek-R1-1.5B & 7308 &  0.2921 & 0.5164 & 0.5896 & 0.4188 \\
        Gemma-2-2B & 9928 & 0.2923 & 0.5191 & \underline{0.5932} & 0.4205 \\
        Phi-2& 10047 & 0.2915 & 0.5166 & 0.5892  & 0.4183 \\
        LLaMA-3.2-3B & 11566 & \textbf{0.2941} & \textbf{0.5205} & \textbf{0.5948} & \textbf{0.4220} \\
        \bottomrule
    \end{tabular}

\end{table}

Model scalability represents a fundamental consideration for practical deployment. We investigate RHYTHM's scaling characteristics across diverse model capacities using pretrained LLMs spanning OPT, LLaMA-3.2, DeepSeek-R1, Gemma-2, Phi-2, and Qwen 2.5 architectures (see~\cref{tab:llm_variants}).
\cref{tab:scale} demonstrates consistent performance gains correlating with increased parameter counts. This scaling relationship necessitates balancing prediction quality against computational requirements. We examine three critical dimensions—predictive accuracy, model size, and per-epoch training duration—as in \cref{fig:scale}. While LLaMA-3.2-3B delivers peak mobility prediction performance, LLaMA-3.2-1B emerges as the pragmatic choice for RHYTHM, optimally balancing accuracy improvements with resource efficiency.

\subsection{Ablation study}
We employ LLaMA-3.2-1B as the standard backbone in all ablation experiments. Component-wise analysis across three datasets (\cref{tab:ablation}) reveals the relative importance of each architectural element. Excluding temporal tokenization causes the most severe degradation at 5.39\%, while removing hierarchical attention (HA) reduces performance by 0.90\%, establishing structured temporal encoding as RHYTHM's fundamental component. For semantic components, both trajectory embeddings and task prompts prove essential, with their joint removal decreasing performance by 1.82\%. Task descriptions demonstrate slightly greater influence, contributing an extra 0.10\% performance loss beyond trajectory information alone.

\begin{table*}[ht]
    \centering
    \caption{\textbf{Ablation study of RHYTHM.} We examine the individual impact of each architectural component on model performance. 
    Results report Accuracy@k metrics with variance below $2\%$. Best performances are marked in \textbf{bold}.
    All components demonstrate substantial contributions to RHYTHM's effectiveness across datasets.}
    \label{tab:ablation}
    \vspace{0.1in}
    \begin{tabular}{l|ccc|ccc|ccc}
        \toprule
        & \multicolumn{3}{c|}{\textbf{Kumamoto}} & \multicolumn{3}{c|}{\textbf{Sapporo}} & \multicolumn{3}{c}{\textbf{Hiroshima}} \\
        \cmidrule(lr){2-4} \cmidrule(lr){5-7} \cmidrule(lr){8-10}
        \textbf{Model} & Acc@1 & Acc@3 & Acc@5 & Acc@1 & Acc@3 & Acc@5 & Acc@1 & Acc@3 & Acc@5 \\
        \midrule
        RHYTHM & \textbf{0.2929} & \textbf{0.5200} & \textbf{0.5835} & \textbf{0.2938} & \textbf{0.4866} & \textbf{0.5502} & \textbf{0.2913} & \textbf{0.5027} & \textbf{0.5753} \\
        w/o HA & 0.2917 & 0.5163 & 0.5881 & 0.2901 & 0.4856 & 0.5481 & 0.2895 & 0.4946 & 0.5657 \\
        w/o token & 0.2801 & 0.5049 & 0.5764 & 0.2768 & 0.4775 & 0.5409 & 0.2749 & 0.4812 & 0.5535 \\
        w/o Traj info. & 0.2914 & 0.5176 & 0.5891 & 0.2879 & 0.4842 & 0.5472 & 0.2858 & 0.4916 & 0.5633 \\
        w/o Task desc. & 0.2895 & 0.5166 & 0.5889 & 0.2883 & 0.4839 & 0.5463 & 0.2882 & 0.4934 & 0.5648 \\
        \bottomrule
    \end{tabular}%

\end{table*}

\end{document}